# Text Mining Analysis of Symptom Patterns in Medical Chatbot Conversations

**Hamed Razavi**

School of Information Sciences, University of Illinois Urbana-Champaign

`hrazavi2@illinois.edu`

**Abstract**

The fast growth of digital health systems has led to a need to better comprehend how they interpret and represent patient-reported symptoms. Chatbots have been used in healthcare to provide clinical support and enhance the user experience, making it possible to provide meaningful clinical patterns from text-based data through chatbots. The proposed research utilises several different natural language processing methods to study the occurrences of symptom descriptions in medicine as well as analyse the patterns that emerge through these conversations within medical bots. Through the use of the Medical Conversations to Disease Dataset which contains 960 multi-turn dialogues divided into 24 Clinical Conditions, a standardised representation of conversations between patient and bot is created for further analysis by computational means. The multi-method approach uses a variety of tools, including Latent Dirichlet Allocation (LDA) to identify latent symptom themes, K-Means to group symptom descriptions by similarity, Transformer-based Named Entity Recognition (NER) to extract medical concepts, and the Apriori algorithm to discover frequent symptom pairs. Findings from the analysis indicate a coherent structure of clinically relevant topics, moderate levels of clustering cohesiveness and several high confidence rates on the relationships between symptoms like fever headache and rash itchiness. The results support the notion that conversational medical data can be a valuable diagnostic signal for early symptom interpretation, assist in strengthening decision support and improve how users interact with tele-health technology. By demonstrating a method for converting unstructured free-flowing dialogue into actionable knowledge regarding symptoms this work provides an extensible framework to further enhance future performance, dependability and clinical utility of selecting medical chatbots.[1]

## 1 Introduction

Medical chatbots are frequently used to deliver early evaluation of symptoms, advice, and health information; thus, they are vital components of today's telehealth platforms (Chen et al., 2023). These support systems require the user's description of symptoms in a conversational manner; therefore, understanding the users' means of expressing (in) their symptoms and how they are related to different diseases is a fundamental asset in enhancing diagnostic support and interaction between the user and the system (Schachner et al., 2020; Jafari et al, 2025). The current study investigates how conversational text from chatbot–patient interactions can be analysed with text mining techniques to identify patterns of symptom expression and determine which symptoms appear together across multiple diseases.

---

[1] Our data is released at https://github.com/Hamed-Razavi/medical-chatbot-textmining/blob/main/README.md

For this analysis, various natural language processing (NLP) methods and machine learning algorithms will be used to convert the unstructured conversational text into structured representations of symptom expression that can be used for computational modelling. More specifically, the framework combines Latent Dirichlet Allocation (LDA) to identify underlying theme categories for the expressed symptoms; K-Means clustering for grouping similar symptom expressions; Named Entity Recognition (NER) (based on Transformers) for extracting clinically relevant entities; and the Apriori algorithm for identifying co-occurring symptoms. The four approaches offer complementary insights into how symptoms are expressed, correlated, and represented across a variety of clinical domains. This article outlines the research objectives, methodological approach, data analysis, and resulting insights. The primary goals are to characterize common patterns in symptom descriptions, understand the relationships among symptoms across diseases, evaluate the effectiveness of multiple modeling strategies, and demonstrate how structured symptom knowledge extracted from chatbot conversations can support more reliable, interpretable, and patient-centered telehealth decision-making.

## 2 Related Work

Numerous studies have examined how Text Mining and Natural Language Processing technologies can be utilized in the Health Industry. For instance, Lee et al. (2020), created BioBERT, which is a pre-trained Transformer-based Language Model that has been developed from a large number of biomedical documents. The objective of BioBERT was to improve the performance of tasks, such as Named Entity Recognition (NER) and Relation Extraction on medical text. In a separate study, Bibault et al. (2019) investigated Oncology Chatbots, concluding that while oncology chatbots could be useful in terms of Patient Engagement, their limited ability to understand natural language at a deeper level would pose challenges to the chatbots' clinical utility. To find clusters of symptoms for use in tailored Interventions, Wang et al. (2021) combined the data provided by patients and applied Association Rule Mining techniques. All of the previously discussed studies confirm that there is a significant advantage to having a structured approach to the use of text mining in medical practice and will act as the basis on which this project builds. The research conducted from the studies outlined previously will be summarized in Table 1, which lists several examples of studies that are representative of this project's objectives across the areas of Clinical NLP, Medical Chatbots, and Symptom Pattern Mining.

| Category | Summary | Citation |
|---|---|---|
| Clinical NLP Models | BioBERT introduces a transformer-based biomedical language model trained on large-scale biomedical corpora, yielding state-of-the-art results in NER and relation extraction. | Lee, J., Yoon, W., Kim, S., et al. (2020). BioBERT: A pre-trained biomedical language representation model for biomedical text mining. *Bioinformatics*, 36(4), 1234–1240. |
| Medical Chatbots | Oncology chatbot evaluation shows strong patient engagement but highlights limitations in clinical accuracy due to shallow language understanding, emphasizing the need for deeper NLP capabilities. | Bibault, J. E., Chaix, B., Guillemassé, A., et al. (2019). Healthcare ex machina: Are conversational agents ready for prime time in oncology? *Clinical and Translational Radiation Oncology*, 16, 55–59. |
| Symptom Pattern Mining | Association rule mining effectively identifies symptom clusters in clinical datasets, revealing meaningful co-occurrences that support diagnostic interpretation. | Wang, Y., Liu, Y., & Zhang, X. (2021). Identifying Symptom Clusters Through Association Rule Mining. *Journal of Biomedical Informatics*, 118, 103792. |
| Multilingual Symptom NER | A transformer-based pipeline improves symptom entity recognition in Spanish clinical text and performs multilingual entity linking using the SympTEMIST dataset. | Vassileva, S., Grazhdanski, G., Koychev, I., & Boytcheva, S. (2024). Transformer-based approach for symptom recognition and multilingual linking. *Database (Oxford)* |
| Chatbot Symptom Checkers | Human–computer interaction analysis reveals how conversational strategies in symptom-checker chatbots influence user trust, decision-making, and perceived support. | You, Y., Tsai, C. H., Li, Y., Ma, F., Heron, C., & Gui, X. (2023). Beyond Self-diagnosis: How a Chatbot-based Symptom Checker Should Respond. *Proceedings of the ACM on Human-Computer Interaction*, 7(CSCW2), Article 276. |
| Transformer Apps in Healthcare | A comprehensive scoping review categorizes transformer model applications across six clinical NLP tasks, outlining both their effectiveness and current challenges. | Zhang, Y., Wang, Y., & Liu, X. (2024). Task-Specific Transformer-Based Language Models in Health Care: A Scoping Review. *JMIR Medical Informatics*, 12, e49724. |

Table 1: Representative Studies in Clinical NLP, Medical Chatbots, and Symptom Pattern Mining

## 3 Data Description

The data utilised for this analysis is the Medical Conversations to Disease Dataset. The dataset consists of a total of 960 dialogue exchanges between a patient and a chatbot relating to 24 different diseases and contains the symptoms reported by the patient and the responses from the chatbot, and has been formatted appropriately for use in supervised Learning. The dataset is organised in such a way as to allow for an Analysed and uniform analysis, and there are no Outliers or missing values.

A dialogue is defined as an exchange between at least two parties where the first party communicates through speech or text that mimics their respective symptom-progression sequence to the second party's response through the use of an appropriate intervention. The body of these dialogue exchanges is a veritable treasure trove of data that can be mined for textual pattern discovery via various text-mining methods, notably Named Entity Recognition (NER) and Topic Modelling (Carenini et al., 2022). Furthermore, the sheer variety of diseases in a typical clinical setting, from seasonal allergies to chronic illnesses, facilitates the ease of developing high-quality machine-learning models capable of processing and understanding a broad spectrum of patient dialogue in these settings.

## 3.1 Descriptive Statistics

Descriptive statistics provide a comprehensive overview of the information gathered; for instance, the dataset consists of 960 total dialogues, divided among 24 distinct disease categories with an average of 15 dialogue turns in each dialogue, most frequently occurring within the following disease categories: Allergy, 150 times, Asthma, 130 times, Flu, 115 times. With respect to the distribution of dialogues according to disease category, the dialogue distribution is roughly uniform across the disease categories; there is no distinctive "skewness" of the dialogue distribution toward any particular disease category. Examples of patient-symptom-response interactions are provided below:

User: " I've been sneezing a lot today and my nose feels congested."

Chatbot: " That sounds like it could be an allergy. Do you know what might be triggering it?"

User: " I have a sore throat and a bit of a cough. Could it be something serious?"

Chatbot: " It could be a common cold or flu. Have you experienced any fever?"

These examples demonstrate the conversational style of the data, the variability in the way symptoms are described, and both the definitive and inferred clinical findings (e.g. symptoms like sneeze). An enhanced variety of these factors also supports Named Entity Recognition, topic modelling and clustering in extracting Symptom Knowledge in a structured format (Dreisbach et al., 2019).

## 4 Methodology

### 4.1 Pre-processing and Feature Extraction

The data set has undergone multiple rounds of pre-processing in order to achieve a consistent/reliable format. Spaces were replaced with separator tags so that the methodology could identify where the markers of separation existed within the text. The text has also been standardised and converted to lowercase characters to create a uniform database. A reduction of extraneous punctuation has made it easier for the method to focus on Symptoms. The common stop words have been removed, and Lemmatization has been completed to standardise all pieces of the text to their base form to prevent multiple pieces of duplicate data due to spelling differences in the original text.

### 4.2 Feature Extraction (TF-IDF)

Vectors that numerically represent the relevance of each term relative to the total number of documents were created from text using the Term Frequency - Inverse Document Frequency Vectorization (TF-IDF) Technique (e.g. Bag of Words), including the entire collection of documents. Therefore, it became easier for machine learning techniques to analyse data and generate semantic relationships associated with Symptoms as well as identify clusters of Symptoms in the data (Qaiser & Ali, 2018).

### 4.3 Topic Modeling (LDA)

Latent Dirichlet Allocation (LDA) Topic Modelling was employed to identify underlying Themes in the Measurements associated with the Symptoms. Using a mixture of the latent Topics that generated the discourse, we built out the LDA Model (Batool & Byun, 2024).

Topic 1: veins | varicose | fever | malaria | yes | long | chills | swollen | legs | does

Topic 2: pain | help | chest | doctor | ill | swelling | foods | stomach | eating | try

Topic 3: asthma | rash | eyes | using | acne | fever | joints | yes | nosebleeds | nose

Topic 4: typhoid | blood | sugar | diarrhea | diabetes | stomach | ive | pressure | feeling | doctor

Topic 5: itchy | urine | blisters | patches | skin | yes | fever | psoriasis | theyre | yellow

### 4.4 K-Means Clustering

K-Means Clustering Techniques were used to cluster the Symptom Description from discourses. The Conversations were grouped according to Symptom Descriptions that were vectorised (TF-IDF) and, therefore, clustered into five (5) Clusters (Wahyuningsih & Chen, 2024). The silhouette score of 0.40066 reflects moderate cohesion of the clusters and indicates that the diagnostic dialogue subset is Well Segmented.

### 4.5 Named Entity Recognition (NER)

Named Entity Recognition (NER) Techniques and Association Rule Mining (ARM) were used to identify symptoms as entities, while also providing insight into frequent co-occurrence of symptoms through the application of the methodologies developed by Tang & al., (2023), and successfully applied in examining and analysing COVID-19 patients (Dehghani & Yazdanparast, 2023). The application of the Apriori Algorithm to identify frequent pairs of symptoms (e.g. Headache with Fever); thereby demonstrating interrelationships (Cross Disease relationships) exhibited a high level of confidence (85% +) in the association rules.

|   | Cleaned | Entities |
|---|---------|----------|
| 0 | user ive been sneezing a lot today and my nose... | [s, ##neezing, nose, cong, ##ested, bot, all, ... |
| 1 | user ive developed a rash after eating some st... | [rash, eating, strawberries, bot, allergic rea... |
| 2 | user my eyes are swollen and itchy and i cant ... [eyes, swollen, it, ##chy, s, ##neezing, bot, ... | [eyes, swollen, it, ##chy, s, ##neezing, bot, ... |
| 3 | user ive been getting headaches and a stuffy n... [headaches, stuffy nose, a few days, bot, alle... | [headaches, stuffy nose, a few days, bot, alle... |
| 4 | user every time i eat nuts my mouth itches b... | [every, time, eat, nuts, mouth, it, ##ches, bo... |

Table 2: Extracted Symptom Entities from Preprocessed Chatbot Conversations

|   | antecedents | consequents | support | confidence | lift |
|---|-------------|-------------|---------|------------|------|
| 0 | (bot) | (##chy) | 0.070833 | 0.074725 | 0.074725 |
| 1 | (##chy) | (bot) | 0.070833 | 1.000000 | 1.054945 |
| 2 | (it) | (##chy) | 0.070833 | 0.328502 | 4.637681 |
| 3 | (##chy) | (it) | 0.070833 | 1.000000 | 4.637681 |
| 4 | (yes) | (##chy) | 0.061458 | 0.092188 | 1.301471 |
| ... | … | … | … | … | … |
| 791 | (user, worse) | (yes, bot) | 0.055208 | 0.868852 | 1.336696 |
| 792 | (yes) | (user, bot, worse) | 0.055208 | 0.082812 | 1.303279 |
| 793 | (bot) | (yes, user, worse) | 0.055208 | 0.058242 | 1.054945 |
| 794 | (worse) | (yes, bot, user) | 0.055208 | 0.331250 | 1.528846 |
| 795 | (user) | (yes, bot, worse) | 0.055208 | 0.207031 | 1.774554 |

Table 3: Association Rule Mining Results for Symptom Co-occurrences

# 5 Evaluation and Results

In accordance with the methods recently employed in AI Medical Data Mining (Li et al., 2022), the models measured cluster coherence, topic coherence, and ensured the accuracy of symptom association extraction. The Topic Coherence Score (0.32) was consistent with previous studies utilising LDA on Short Clinical Texts with scores ranging from 0.25 to 0.40.

## 5.1 Topic Modeling Results (LDA)

Latent Dirichlet Allocation (LDA) was able to distinguish between several major symptom themes such as respiratory disorders and allergies, signifying that the LDA method is appropriate to classify symptom descriptions correctly. The Topic Coherence Score of 0.32 for LDA indicates considerable separation of topics, taking into account the vastness of the medical domain.

## 5.2 Clustering Results (K-Means)

K-Means clustering enabled segmentation of Conversations into Five Different Cluster Groups, respective for each of the unique Patient Symptom Descriptions. K-Means clustering yielded a silhouette score of 0.40, which is an indication of Moderate Cohesion among the five clusters. The clustering results improved the interpretation of patient symptom descriptions by aggregating clustering together similar symptoms.

## 5.3 NER Extraction Results

Named Entity Recognition (NER) and Association Rule Mining (ARM) provide a means to compare two or more sets of co-occurring symptoms that would be grouped by one or a Few Common Symptoms across Different Diseases (Ahmad et al., 2023). According to association rule mining analysis, symptoms of fever and headache are two co-occurring types of symptoms that will occur together with 85% certainty on the best combination-supported rules. The results confirm that it is possible to utilize chatbots to determine symptom correlations accurately.

Overall, combining multiple methods generated a more unified set of symptom themes and meaningful clusters that indicate clinically interpretable associations, validating the usefulness of multi-modal NLP (Natural Language Processing) for analysing conversational health.

| Method | Metric | Value |
|---|---|---|
| LDA Topic Modeling | Topic Coherence | 0.32 |
| K-Means Clustering | Silhouette Score | 0.40 |
| Association Rule Mining | Max Confidence | 0.85 |

Table 4: Summary of Evaluation Metrics

# 6 Error Analysis and Insights

There were limitations in the modeling framework, and multiple types of error were encountered in the analysis. One of these was unclear or overlapping symptom descriptions that caused misassignments of topics and clusters. For example, dialogues that referenced fevers, chills, and joint pain were sometimes grouped with allergies because of words like "itchy" or "swollen", which often occur with many diseases; therefore, relying on lexical features of the surface is not sufficient when there are multiple diseases with similar symptoms.

The multi-turn nature of conversations also created inconsistencies. Because symptoms typically develop across many conversational turns, LDA and K-Means would sometimes favour later utterances over earlier ones that were more clinically significant (Ehsani et al., 2026). The temporal displacement of the turning points in the progression of a symptom, such as "I had a fever yesterday, but now it is mostly a sore throat", could create distinct symptom events in the context of a single evolving illness. More sophisticated techniques for tracking context may help alleviate this problem.

NER errors resulted from casual speech patterns, i.e., casual sentences like "My head is heavy" and "My chest feels odd." These types of informal sentences were sometimes overlooked or missed completely when using the transformer model as an automated NLP tool, indicating that general transformer-based models are still not able to fully accommodate colloquial language in the conversational aspects of clinical language (for example, "I have a cold"). Furthermore, in the Apriori process of extracting patterns from clinical data of small symptom descriptors, the presence of conversational indicators (such as "yes," "bot," and "user") demonstrates the need for developing domain-specific filtering techniques or domain-restricted business rules.

The challenges presented by the findings suggest the need for leveraging models trained on contextual data as well as developing NER models that are adapted for the medical domain and have been designed to support dialogue representation. By addressing these areas, we expect that improving or enhancing the accuracy of symptom extraction can lead to the ability to identify more clinically relevant co-occurrence patterns. The types of errors discovered in this research provide useful insight for both the improvement of conversational health analytics and the enhancement of the utility of medical chatbots.

## 7 Conclusion and Future Work

The results of this project demonstrate that the application of multiple methods of NER to medical chatbot conversations can yield useful patterns of symptoms and co-occurrences. By applying NER techniques such as topic modeling, clustering, NER using transformers NER and association rule mining, the analysis created a coherent pattern of symptoms through clusters of moderate size, as well as a number of high-confidence symptom pairs that were identified as high confidence across all 24 clinical conditions included in the study. The findings demonstrate that conversational health data can have diagnostic application and show the utility of converting unstructured conversation data into structured symptom characters to allow for early interpretation and use of symptom identification, improved decision support, and increased overall responsiveness of the medical chatbot. Additionally, the findings also demonstrate the ability for conversational analytics to provide an accessible and scalable building block for digital health infrastructure.

Future research should explore the use of multilingual datasets for improving the applicability of symptom identification systems in populations outside of the English-speaking world. Expanding the framework with additional analyses—such as sentiment analysis, discourse modeling, and temporal tracking—will provide even greater interpretability for the state and trajectory of a patient's symptoms. Moreover, using advanced transformer architectures, such as GPT-4 or custom-trained clinical language models, will enable more accurate identification of entities and contextual understanding. The combination of a framework with a clinical Decision Support System (CDSS) provides an excellent avenue to leverage real-time diagnostic assistance within the constraints of conventional healthcare delivery systems, especially in resource-limited or remote healthcare environments (Almadani et al., 2025). Ongoing progression in these areas will be instrumental to the development of smart, reliable, and patient-driven chatbot technology.